\documentclass[runningheads]{llncs}
\usepackage{multirow}
\usepackage{graphicx}
\usepackage{subcaption}
\usepackage{amsmath}
\usepackage{lipsum}
\usepackage{hyperref}

\newcommand{\xnot}{{\sf xNot360}}

\begin{document}
\title{Balancing Exploration and Exploitation in LLM using Soft RLLF for Enhanced Negation Understanding}

\titlerunning{ }

\author{Ha-Thanh Nguyen \and
Ken Satoh }
\authorrunning{ }
%
\institute{National Institute of Informatics, Tokyo, Japan \\ \email{\{nguyenhathanh,ksatoh\}@nii.ac.jp}
}

\maketitle              
\begin{abstract}
Finetuning approaches in NLP often focus on exploitation rather than exploration, which may lead to suboptimal models. Given the vast search space of natural language, this limited exploration can restrict their performance in complex, high-stakes domains, where accurate negation understanding and logical reasoning abilities are crucial.  To address this issue, we leverage Reinforcement Learning from Logical Feedback (RLLF) to create an effective balance between exploration and exploitation in LLMs. Our approach employs an appropriate benchmark dataset for training and evaluation, highlighting the importance of exploration in enhancing negation understanding capabilities. We compare the performance of our RLLF-enhanced LLMs with baseline models trained without RLLF, demonstrating the value of this balanced approach. Furthermore, we showcase the potential of our method in legal AI applications by employing transfer learning and evaluating its impact on negation understanding. Our experimental results exhibit the effectiveness of balancing exploration and exploitation with RLLF in improving LLMs' negation capabilities. This has implications for the development of more accurate, reliable, and logically consistent language models in high-stakes domains.

\end{abstract}

\section{Introduction}
Negation, a fundamental aspect of communication in natural language, remains a challenging concept for state-of-the-art pre-trained language models like GPTs \cite{radford2019language,brown2020language,ouyang2022training,openai2023gpt4}. Outstanding in various tasks, these models often exhibit shortcomings when grappling with the intricacies of negation, an issue that becomes particularly pronounced in the context of legal AI, where precision in understanding is non-negotiable \cite{kim2019statute,hossain2020analysis,geiger2020neural,hosseini2021understanding,hossain2022analysis,nguyen2023negation,nay2023large}. The complexity of negation requires a deep understanding and accurate computational treatment, which is critical in high-stakes domains such as law.

As the field of natural language processing (NLP) evolves, the treatment and comprehension of negation in textual data stand as a testament to a model's sophistication and practical application, especially in legal settings. Addressing this, initiatives have surfaced to enhance language models' capabilities through various means, including the augmentation of models with negation-focused datasets like MoNLI \cite{geiger2020neural}, CondaQA \cite{ravichander2022condaqa}, Jina \cite{gunther2023jina} and unlikelihood training objectives \cite{hosseini2021understanding}. While these initiatives have shown progress in enhancing language models' capabilities in handling negation, fully addressing the nuanced challenges presented by negation remains an ongoing task.

\begin{figure}[h]
  \centering
  \includegraphics[width=0.8\textwidth]{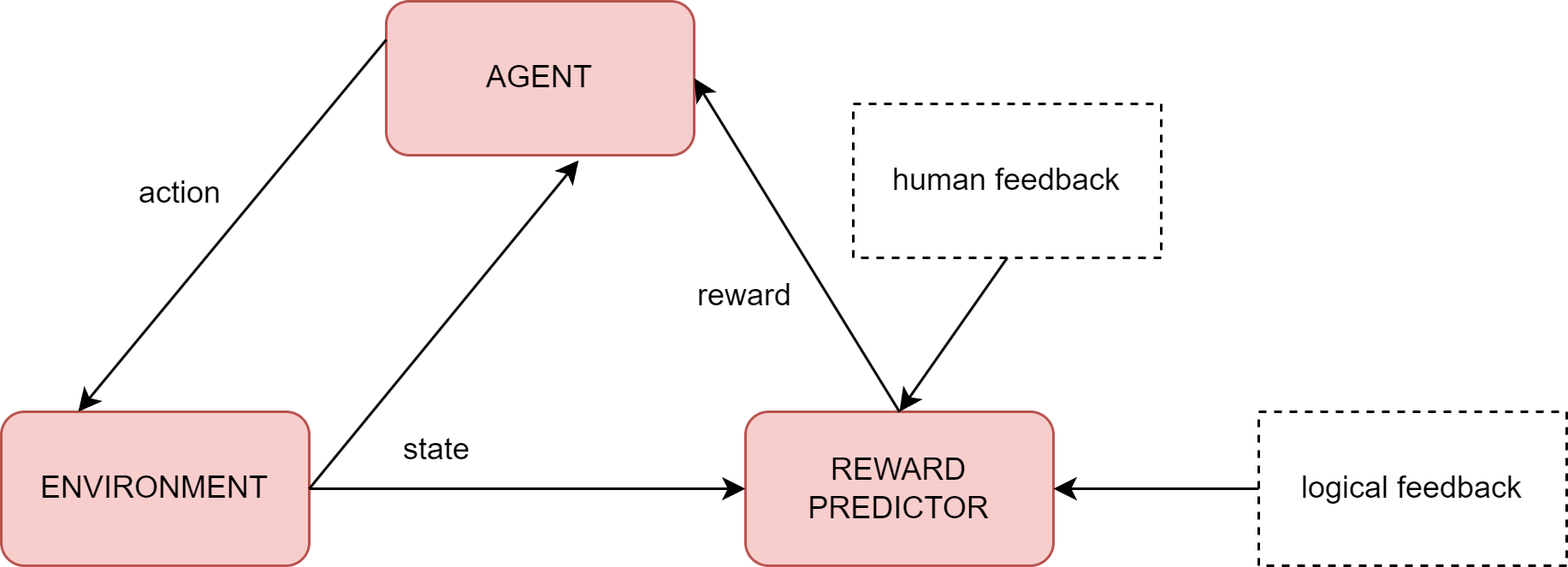}
  \caption{RLLF is the idea of allowing feedback for reinforcement learning to come not only from the user but also from the accuracy in the chain of logical reasoning. \cite{nguyen2023enhancing}}
  \label{fig:rllf}
\end{figure}

Advances in language model development have seen the introduction of reinforcement learning from human feedback (RLHF) \cite{ouyang2022training}, which, while effective in aligning models with human intent, simultaneously exposes the limitations of human bias and the necessity for logical coherence \cite{nguyen2023legal}. In this context, the legal domain emerges as a pivotal use case that requires a blend of factual accuracy and stringent logical reasoning  — competencies that are indispensable for reliable decision-making.
In contrast to the reliance on human feedback in RLHF, Reinforcement Learning from Logical Feedback (RLLF) \cite{nguyen2023enhancing} allows the reward predictor to be trained using logical feedback in addition to human evaluations. This method aims to enhance LLMs' logical reasoning capabilities while minimizing the impact of human biases.

In this paper, we propose and experiment with a method called Soft RLLF, which uses logical signals from natural language rather than logic programming languages. This choice is a practical decision for experimentation, as it considers the challenges associated with closed models that do not allow reinforcement learning and the high costs of working with models containing billions of parameters. Moreover, it takes into account the feasibility of working with smaller models in a laboratory setting, acknowledging that higher levels of logic programming capabilities may be more suited for larger commercial models. The experimental goal of this paper is not to create a state-of-the-art (SOTA) performance model but to demonstrate the effectiveness of the proposal and provide insights into its potential benefits. By using Soft RLLF, we aim to improve model performance through increased exploration of the vast search space of natural language.

\section{Background}
Generative Pre-trained Transformer 2 (GPT-2) is an advanced language model developed by OpenAI \cite{radford2019language}. As the successor of GPT-1, GPT-2 marks a significant leap in language modeling by increasing the parameter count tenfold and enhancing the training dataset size proportionally \cite{zhu2015aligning}. Unlike previous models that employed recurrence-based or convolution-based architectures, GPT-2 utilizes a transformer-based approach which relies on attention mechanisms \cite{vaswani2017attention}. This allows the model to focus selectively on relevant segments of input text through effective parallelization, outperforming existing benchmarks.
Despite its achievements, GPT-2 also exhibits limitations, particularly when generating longer texts, which can become repetitive and nonsensical \cite{radford2019language}.

Throughout the development of the GPTs, the automated understanding of language has evolved considerably. GPT-2 \cite{radford2019language} demonstrated impressive zero-shot learning capabilities on language modeling tasks. GPT-3 \cite{brown2020language}, with its increased model size, achieved strong performance on numerous NLP tasks, including translation and question-answering. The GPT-3.5 model (a.k.a. ChatGPT) \cite{ouyang2022training} introduced a method labeled as reinforcement learning from aggregate human feedback (RLHF) to better align the model with user intent. However, this approach has led to sacrificing the model's logical abilities in exchange for user satisfaction.

Table \ref{tab:model_performance} summarizes the performance of the GPT models on  \xnot{} dataset in the zero-shot setting. The \textit{eXploring Negation Over Text with 360 samples} (\xnot{}) dataset \cite{nguyen2023negation} was developed to explicitly assess the negation detection abilities of language models, such as GPT-2, GPT-3, GPT-3.5, and GPT-4. The dataset was designed to take into account a wide range of sentence pairs containing diverse negation structures and various language contexts. By creating a more challenging benchmark for negation detection, \xnot{} aims to further understand the limitations of current state-of-the-art pre-trained language models and offer valuable insights into potential improvements in natural language understanding.

In the table, we can see that GPT-4 outperforms the other models in both major metrics, achieving an accuracy of 0.7833, an F1-score of 0.7706, a precision of 0.8187, and a recall of 0.7278. On the other hand, GPT-3.5 shows a significant performance dip compared to its counterparts, with an accuracy of 0.4306, an F1-score of 0.2705, a precision of 0.3762, and a recall of 0.2111. GPT-3 exhibits moderate performance, with an accuracy of 0.6056, an F1-score of 0.6913, a precision of 0.5679, and a recall of 0.8833. Lastly, GPT-2 has the lowest accuracy (0.5000) but achieves the highest recall (1.0000). This demonstrates that the model classified all samples as positive, ensuring that no positive samples were omitted \cite{nguyen2023negation}.

\begin{table}
\centering
\footnotesize
\begin{tabular}{|c|c|c|c|c|}
\hline
\textbf{Model} & \textbf{Accuracy} & \textbf{F1-score} & \textbf{Precision} & \textbf{Recall} \\ \hline
GPT-2           & 0.5000            & 0.6667            & 0.5000             & 1.0000           \\ \hline
GPT-3           & 0.6056            & 0.6913            & 0.5679             & 0.8833           \\ \hline
GPT-3.5         & 0.4306            & 0.2705            & 0.3762             & 0.2111           \\ \hline
GPT-4           & 0.7833            & 0.7706            & 0.8187             & 0.7278           \\ \hline
\end{tabular}
\caption{Performance comparison of GPT models on the \xnot{} dataset.\cite{nguyen2023negation}}
\label{tab:model_performance}
\end{table}

\begin{figure}
\centering
\includegraphics[width=.75\textwidth]{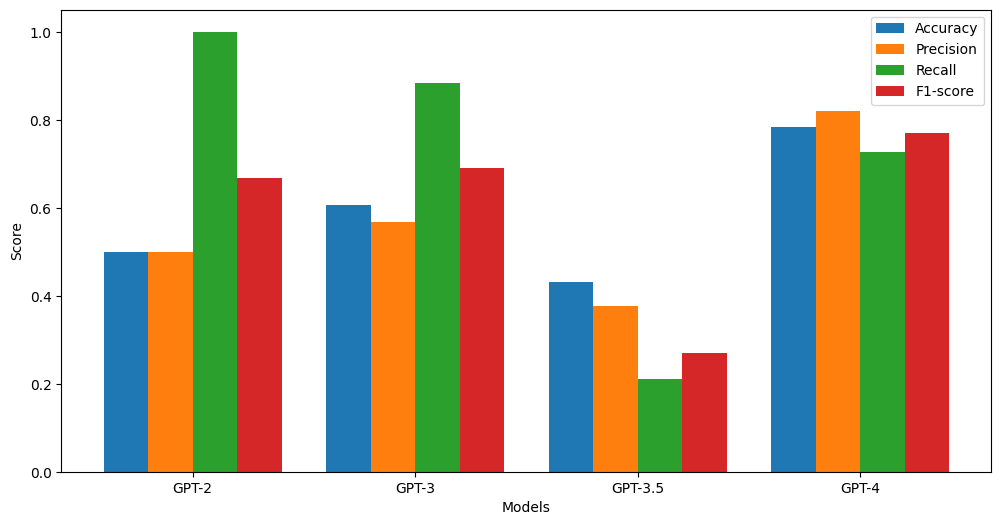}
\caption{Performance chart of GPT models on the \xnot{} dataset. The chart displays a sinusoidal-like pattern, highlighting the differences in performance among the models.\cite{nguyen2023negation}}
\label{fig:performance_chart}
\end{figure}

To provide a visual representation of the performance variations among the four GPT models, Figure \ref{fig:performance_chart} plots a performance chart that illustrates the fluctuating nature of their accuracy, F1-score, precision, and recall metrics. The chart exhibits a sinusoidal-like pattern, indicating the disparities in performance across the models.

As shown in Figure \ref{fig:performance_chart}, the performance of the GPT models varies considerably. GPT-4 exhibits the highest performance across major metrics, while GPT-3.5 demonstrates a noticeable dip in performance. GPT-3, on the other hand, shows moderate performance and GPT-2 has the lowest accuracy but the highest recall. From GPT-3.5, RLHF is applied. This observation suggests that optimizing parameters to align with user preferences may compromise the model's logical reasoning abilities, especially for models with limited size. With larger models like GPT-4, this issue is partially addressed, but there remains substantial room for improvement in terms of performance.

In high-stakes domains like law, and healthcare, prioritizing adherence to logical rules over satisfying all user requirements is deemed more important, incorporating logical feedback into the model would be necessary. This concept is the driving force behind reinforcement learning with logical feedback – an approach to refine the model's performance by integrating explicit feedback on logical reasoning.


Reinforcement Learning from Logical Feedback (RLLF) \cite{nguyen2023enhancing} is an approach that focuses on addressing the limitations of Reinforcement Learning from Human Feedback (RLHF), which is susceptible to biases introduced by human feedback and the need for complex logical reasoning, especially in high-stakes domains. By using logical feedback in addition to human evaluations, the RLLF method aims to enhance the logical reasoning capabilities of language models, striking a balance between ensuring both user satisfaction and logical accuracy.

By employing the RLLF framework, language models can enhance their logical reasoning capabilities while minimizing the influence of human biases and subjective feedback. This innovative approach is particularly suitable for logic-intensive domains, where the ability to reason logically and provide accurate information is essential for practical applications and decision-making processes. The logic verifier can be adapted to different logical frameworks and languages, making the RLLF approach versatile and applicable to a wide range of scenarios.

\section{Methodology}

\begin{figure}
\centering
\includegraphics[width=.9\textwidth]{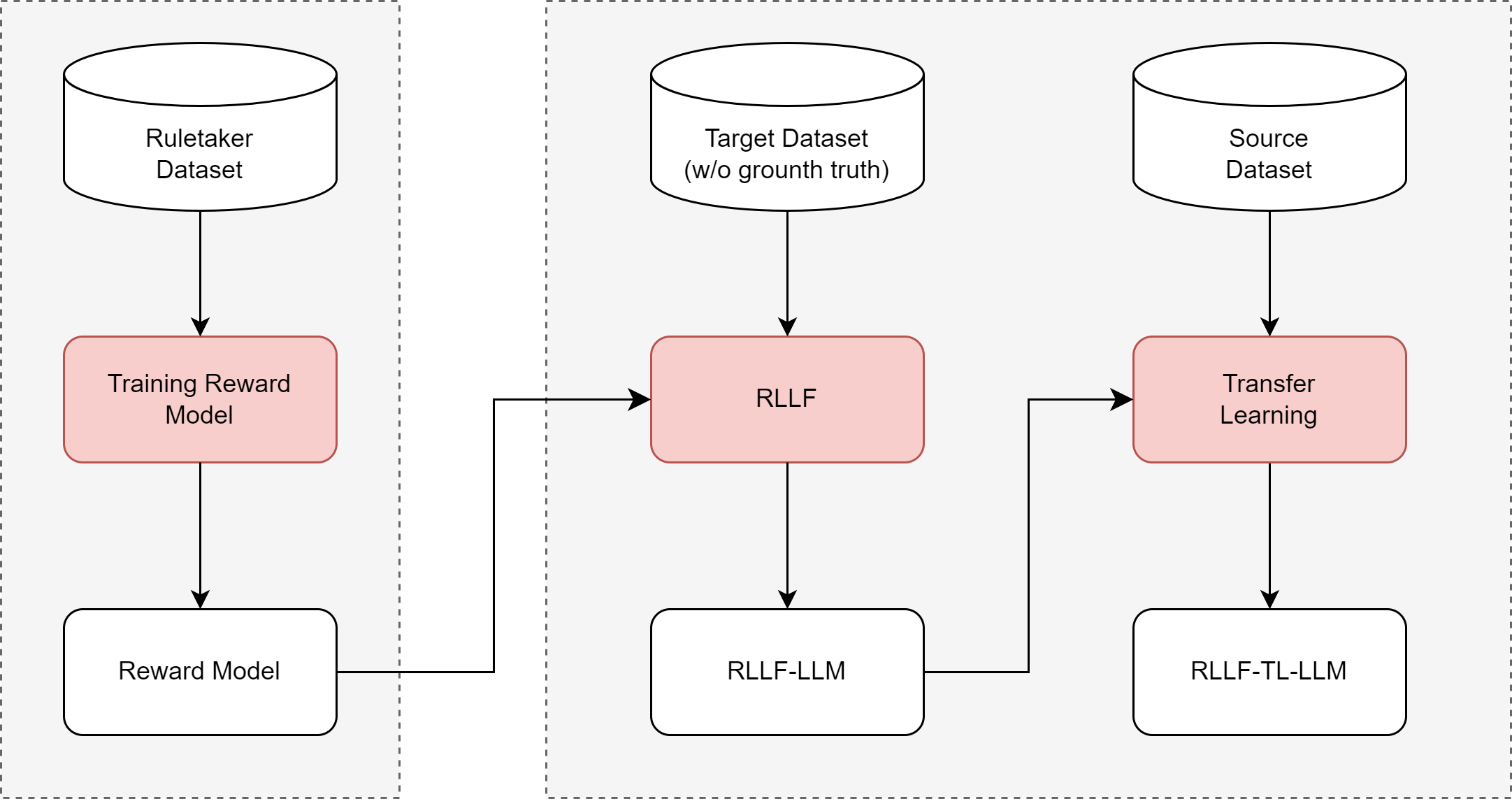}
\caption{Overview of the Reinforcement Learning from Logical Feedback (RLLF) methodology used to improve LLM's negation understanding capabilities, highlighting the exploration-exploitation balance and key steps involved.}
\label{fig:method}
\end{figure}

In this section, we present our approach to enhancing LLM's negation understanding capabilities by employing \textit{Soft} RLLF, focusing on the balance between exploration and exploitation. The methodology consists of three steps as demonstrated in Figure \ref{fig:method}, including \textit{Training Reward Model}, \textit{RLLF} and \textit{Transfer Learning}.

Our approach employs RLLF as a means to supplement LLM's exploration ability. In the context of machine learning, exploration and exploitation refer to two fundamental strategies that a model can use to interact with its environment and learn from data.
Exploration involves observing new, unseen samples from the environment, allowing the model to expand its knowledge about various possible scenarios and potentially improve its decision-making process. A higher degree of exploration can lead to improved generalizability of the model as it helps the model understand more diverse situations and contexts.

With RLLF, we aim to improve the model's negation understanding by encouraging it to explore a broader range of negation possibilities during training. The key idea is to generate negated sentences from LLM itself and verify the quality of these generated sentences through a reward model. While exploitation is necessary for ensuring that the model can make accurate predictions and decisions, relying solely on exploitation can lead to overfitting and reduced generalizability. Striking an appropriate balance between exploration and exploitation is essential for enhancing the overall performance, robustness, and generalizability of the model.

\subsection{Training Reward Model}

The first step in our methodology involves training a reward model using a large dataset centered around logical reasoning. To achieve this, we employ supervised learning by feeding the model a logically-focused dataset containing a broad range of sentence structures and logical contexts. Let the dataset be represented as $D = \{(x_i, y_i)\}_{i=1}^{N}$, where $x_i$ is the input sentence, $y_i \in \{0, 1\}$ is the binary label indicating the logical relationship, and $N$ is the total number of instances in the dataset.

The objective of the training process is to learn a reward model $R$ that maps an input sentence $x_i$ to its corresponding logical label $y_i$. In other words, we want to find the optimal model parameters $\theta^*$ that minimize the loss function $L$:

\begin{equation}
    \theta^* = \arg\min_\theta L(\theta, D) = \arg\min_\theta \sum_{i=1}^{N} l(R_\theta(x_i), y_i),
\end{equation}

where $l$ is the cross-entropy loss representing the divergence between the model's predictions and the true labels. To optimize the model parameters $\theta$, we can employ gradient-based optimization techniques, such as stochastic gradient descent (SGD) or Adam.


\subsection{Reinforcement Learning on Logical Feedback}
We apply Soft RLLF to the LLM on the target dataset using the reward model as the logical verifier. Soft RLLF, in this context, is an adaptation of the RLLF approach that utilizes natural language signals instead of logic programming languages to provide logical feedback during the training process. For each sentence in the target dataset, the LLM generates a negated sentence without knowledge of the ground truth label, and the reward model evaluates the generated negated sentence based on its confidence level of negation with the original sentence:

$$
R(S_i, NS_i) = L(S_i, NS_i),
$$

where $R$ is the reward signal, $S_i$ is the original sentence from the target dataset, $NS_i$ is the negated sentence generated by the LLM, and $L$ is the logical evaluation score provided by the Reward Model on a scale of 0 to 1, with 1 being the highest and 0 being the lowest.

The LLM is updated using the reward signal from the Reward Model, optimizing its parameter weights to maximize the expected reward:

$$
\theta_{t+1} = \theta_{t} + \alpha R(S_i, NS_i) \nabla_\theta \log P_\theta(NS_i \mid S_i),
$$

where $\theta_{t}$ and $\theta_{t+1}$ denote the model parameters at time $t$ and $t+1$, respectively, $\alpha$ is the learning rate, and $P_\theta$ is the conditional probability given by the LLM for the generated negated sentence $NS_i$ given the original sentence $S_i$.

By following these steps, the LLM incorporates exploration during its training process by generating and evaluating a diverse range of negated sentences, ultimately improving its negation understanding capabilities.

\subsection{Transfer Learning and Model Evaluation}
To demonstrate the effectiveness of our RLLF-enhanced exploration approach for improving LLM's negation understanding capabilities, we perform transfer learning using text classification with LLM as the backbone. We compare the results of two different settings:

\begin{enumerate}
    \item \textbf{Without RLLF}: A baseline LLM model is trained using supervised training without RLLF enhancement.
    \item \textbf{With RLLF}: LLM is trained using our RLLF-enhanced exploration approach as described earlier.
\end{enumerate}

For each setting, we train a text classification model using a source dataset in the near domain and evaluate its performance on the target test set. We compare the accuracy, precision, recall, and F1-score for the two settings to assess the benefits of incorporating RLLF-enhanced exploration into the LLM model training process.

The comparison between the two settings allows us to observe how incorporating RLLF-enhanced exploration during the training stage affects the model's performance when applied to a real-world domain, such as the legal domain. Our hypothesis is that the RLLF-enhanced LLM would exhibit a more accurate understanding of negation compared to the baseline model, as the RLLF approach specifically focuses on enhancing the model's logical reasoning capabilities using a reward signal based on logic consistency.

By using transfer learning, we effectively leverage the pre-trained LLM models adapted to different training settings and extend their capabilities in a broader scope. The evaluation of the classification performance of these models, as measured by the accuracy, precision, recall, and F1-score, helps us quantify the effectiveness of the RLLF-enhanced exploration approach for improving negation understanding capabilities in comparison to traditional supervised training without RLLF.

\section{Experiment}

In this section, we describe the experimental setup to evaluate the benefits of incorporating RLLF-enhanced exploration into the LLM training process. 

\subsection{Model Choice}
Given the different versions of GPT models and other LLMs such as LLaMA~\cite{touvron2023llama}, Bard\footnote{\url{https://bard.google.com/}}, Jurassic-2\footnote{\url{https://www.ai21.com/blog/introducing-j2}}, and Claude\footnote{\url{https://www.anthropic.com/index/introducing-claude}}, choosing an appropriate model for experimentation is crucial. We opted for GPT-2 as the primary model in this study for several reasons. First, GPT-2 offers cost-efficiency compared to larger models with billions of parameters, making it more accessible for research purposes. Second, GPT-2 allows for considerable customization, permitting researchers to finetune, adapt, and modify the model to fit their experimental requirements.

Basing our experiments on GPT-2 enables us to draw conclusions from a myriad of experiments across different scales and settings. While the findings based on GPT-2 may not directly mirror the results from more advanced models or the largest LLMs, the insights gleaned from our study can provide valuable guidance for further research. By examining the strengths, weaknesses, and opportunities for improvement in negation understanding abilities in GPT-2, we pave the way for future research on larger and more advanced LLMs.

Table \ref{tab:model_performance} shows the performance of GPT-2 on the \xnot{} dataset in a zero-shot setting. Although not as poor as GPT-3.5, which attempts to answer and achieves an accuracy below random guess, GPT-2 also demonstrates that the zero-shot setting is not sufficient for the model to provide useful information. Since \xnot{} is a dataset designed for binary classification, if a model consistently selects one label, it would achieve 50\% accuracy without needing any additional skills. However, these results serve as a good starting point for us to verify our hypothesis regarding the effectiveness of RLLF and transfer learning.


\subsection{Datasets}

\subsubsection{Ruletaker Dataset \cite{ruletaker2020}:}

The Ruletaker dataset is employed to train the reward model. Designed by AllenAI, this dataset aims to train transformers to reason over language or emulate reasoning, with a focus on chains of reasoning, including deductive inference in constrained settings. The dataset contains 480,000 training samples, 76,000 development samples, and 150,000 test samples.
The Ruletaker dataset is designed to test reasoning over synthetically generated natural language sentences, bypassing the need for a formal representation. 

The large size and diverse sentence structures and logical contexts of the Ruletaker dataset make it suitable for training a reward model focused on logical reasoning. By incorporating this dataset in the reward model training process, we can enhance the logical reasoning capabilities of the GPT-2 model, with potential applications in high-stakes domains such as legal AI.

\subsubsection{Jina Dataset \cite{gunther2023jina}:}

The Jina dataset is used for performing transfer learning with 10,000 samples. As demonstrated by our analysis of the GPT-2 model, the zero-shot setting is insufficient for smaller LLMs like GPT-2 to be useful in negation detection problems. We use the Jina dataset as training data within the near domain of transfer learning.

This dataset consolidates various retrieval objectives, such as e-commerce search, duplicate detection, web retrieval, article retrieval for question-answering, and text classification. By reformulating this dataset into sentence pairs with 0 and 1 labels, we can indicate the presence or absence of negation in the meaning of the pair, respectively. This reformatted dataset enables us to analyze the performance of GPT-2 in the context of negation understanding through transfer learning.

\subsubsection{\xnot{} Dataset \cite{nguyen2023negation}:}

The \xnot{} dataset serves two primary purposes: (1) evaluating the performance of GPT-2 in different settings, and (2) providing queries for the GPT-2 model during the RLLF process. The use of this dataset enables GPT-2 to explore negation in sentence pairs without being biased by any specific patterns. Generated negation sentences are then evaluated by the reward model, providing logical feedback to update the GPT-2 model.

The \xnot{} dataset is designed to test the logical understanding of LLMs through the subtlety of natural language with 360 samples. In some cases, the word ``not'' does not carry a negation meaning. Instead, it requires a deeper understanding of the logical concepts underlying the language to correctly determine the presence or absence of negation in a given sentence. 

Table \ref{tab:truth_value} demonstrates the complexity of negation in natural language, showing an instance where the logical expressions of two related sentences differ.

\begin{table}[h]
\centering
\footnotesize
\begin{tabular}{|c|c|c|c|c|}
\hline
A & B & $\lnot A \lor B$ & $A \lor \lnot B$ & $\lnot A \lor B \neq A \lor \lnot B$ \\
\hline
T & T & T & T & F \\
T & F & F & T & T \\
F & T & T & F & T \\
F & F & T & T & F \\
\hline
\end{tabular}
\caption{Truth table illustrating the complexity of negation in natural language. \cite{nguyen2023negation} \label{tab:truth_value}}
\end{table}

\begin{table}
  \centering
  \begin{tabular}{|l|c|c|c|}
    \hline
    \textbf{Step} & \textbf{Hyperparameter} & \textbf{Value} \\
    \hline
    \multirow{5}{*}{Train Reward Model} 
    & Learning Rate & 2e-5 \\
    & Train Batch Size & 48 \\
    & Eval Batch Size & 48 \\
    & Num. Train Epochs & 1 \\
    & Weight Decay & 0.01 \\
    \hline
    \multirow{3}{*}{GPT in RLLF} 
    & Min\_length & -1 \\
    & Top\_k        & 0.0 \\
    & Top\_p        & 1.0 \\
    \hline
    \multirow{2}{*}{PPO in RLLF} 
    & Learning Rate & 1e-5 \\
    & Batch Size   & 16 \\
    \hline
    Transfer Learning & Learning Rate & 1e-5 \\
    \hline
  \end{tabular}
  \caption{Hyperparameters for the experimental steps, determined through random search.}
  \label{tab:hyperparameters}
\end{table}

The use of the \xnot{} dataset in our experiments enables a more effective evaluation of the GPT-2 model's performance in negation detection and enhances our understanding of the capabilities of LLMs in this context. By employing this dataset, we can identify areas for improvement in the GPT-2 model's understanding of negation and better assess the effectiveness of our RLLF-enhanced exploration and transfer learning approaches.

\subsection{Experimental Settings}

The experiment consists of the following steps:

\begin{enumerate}
    \item \textbf{Train reward model - RuletakerBert}: Train the reward model using the Ruletaker dataset to serve as a logical verifier in the RLLF training process. We use bert-base-cased as the core model for this item. With a dataset of 480,000 training samples, we found that 1 epoch is enough for the model to achieve good accuracy, 0.9312 is the performance on the test set.
    \item \textbf{RLLF training - GPT-2-RLLF}: Perform RLLF training on GPT-2 using the \xnot{} dataset as the target dataset and RuletakerBert as the logical verifier. This process stops when there is no improvement in the reward score. The resulting model, RLLF-GPT-2, incorporates the benefits of RLLF-enhanced exploration. 
    \item \textbf{Transfer learning - GPT-2-RLLF-TT}: Apply transfer learning to RLLF-GPT-2 using the Jina Negation Dataset as the source dataset, yielding the RLLF-TT-GPT-2 model. We use an early stop setting, the training process stops when there is no improvement on the evaluation set of the Jina dataset.
    \item \textbf{Model evaluation}: Evaluate the performance of the final GPT-2-RLLF-TT model on the \xnot{} dataset with ground truth. Compare accuracy, precision, recall, and F1-score with those of a baseline GPT-2 model trained using supervised learning without RLLF enhancement (GPT-2-TT) and GPT-2 with zero-shot setting (GPT-2-ZS).
\end{enumerate}

In the training process, we use A100 GPUs. The batch size values mentioned for both training and evaluation parts take this hardware into account for efficient utilization. The chosen hyperparameters for the training processes, determined through random search, are summarized in Table \ref{tab:hyperparameters}.

\subsection{Experimental Results}

\begin{figure*}[h]
\centering
\begin{subfigure}{.3\textwidth}
  \centering
  \includegraphics[width=\linewidth]{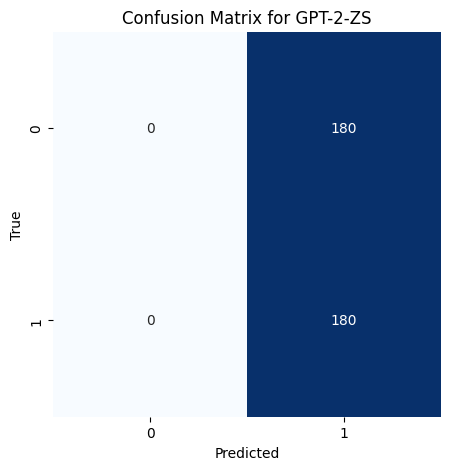}
  \caption{GPT-2-ZS}
  \label{fig:gpt2_zs}
\end{subfigure}
\begin{subfigure}{.3\textwidth}
  \centering
  \includegraphics[width=\linewidth]{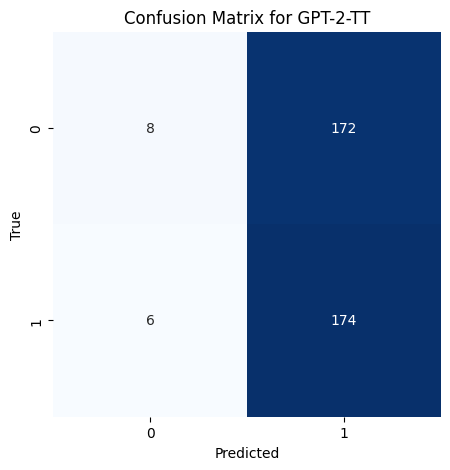}
  \caption{GPT-2-TT}
  \label{fig:gpt2_tt}
\end{subfigure}
\begin{subfigure}{.3\textwidth}
  \centering
  \includegraphics[width=\linewidth]{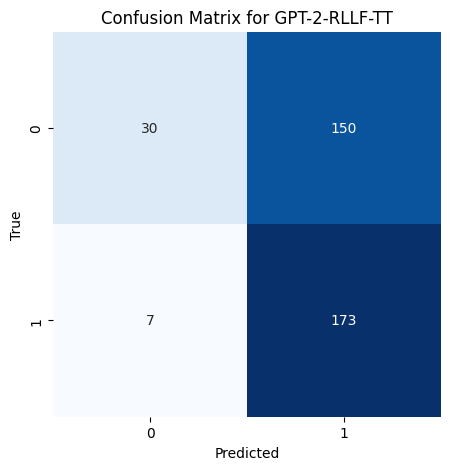}
  \caption{GPT-2-RLLF-TT}
  \label{fig:gpt2_rllf_tt}
\end{subfigure}

\caption{Confusion matrices of GPT-2-ZS, GPT-2-TT, and GPT-2-RLLF-TT predictions on \xnot{}.}
\label{fig:confusion_matrices}
\end{figure*}

The accuracy rates of the three models are shown in Table \ref{tab:comparison}, their confusion matrices are display in Figure \ref{fig:confusion_matrices}.

The GPT-2 model in the zero-shot setting has an accuracy rate of 50\%. As discussed earlier, GPT-2-ZS demonstrates a poor understanding of negation and always predicts the same label (i.e., no negation detected). The model's performance is equivalent to random chance, providing unreliable results for enhancing the understanding of negation.

The GPT-2 model with transfer learning without RLLF has an accuracy rate of 50.56\%. Although it exhibits a slight improvement over GPT-2-ZS, the performance of GPT-2-TT can be considered as marginally above random guessing. The inability to achieve dramatically improved accuracy highlights the limitations of solely relying on traditional supervised learning and not incorporating exploration-enhancing methods, like RLLF.

The GPT-2 model with RLLF-enhanced exploration and transfer learning attains an accuracy rate of 56.67\%. As the best-performing model among the three, GPT-2-RLLF-TT shows the benefits of using RLLF and transfer learning for improving negation understanding. The improvement in accuracy, while significant, is not yet optimal, highlighting the need for further enhancements and exploration of the model's abilities to address negation understanding in a more accurate and comprehensive manner.

The results show that the combination of RLLF-enhanced exploration and transfer learning in the GPT-2-RLLF-TT model can lead to substantial improvements in accuracy. This outcome underscores the importance of balancing exploration and exploitation in LLMs to enhance their negation understanding capabilities more effectively.

\begin{table}
\centering
\begin{tabular}{|l|c|}
\hline
\textbf{Model} & \textbf{Accuracy} \\ \hline
GPT2          & 0.5000            \\ \hline
GPT2-TT       & 0.5056            \\ \hline
GPT2-RLLF     & 0.5667            \\ \hline
\end{tabular}
\caption{Performance of GPT-2 with different settings on \xnot{} dataset.}
\label{tab:comparison}
\end{table}

By adopting methods like RLLF-enhanced exploration and transfer learning, researchers can further improve LLMs' performance, paving the way for the development of more accurate, robust, and logically consistent language models in high-stakes domains where accurate negation understanding is vital to decision-making processes.

\section{Discussion}

In this study, we demonstrate the potential of balancing exploration and exploitation in LLMs using RLLF to enhance their negation understanding capabilities. Our experimental findings show that incorporating RLLF-enhanced exploration and transfer learning into GPT-2 training leads to improved performance, as evidenced by the increased accuracy rate of the GPT-2-RLLF-TT model. This result highlights the benefits of using RLLF and transfer learning for promoting LLM's negation understanding.

However, our research also exposes several limitations. First, a significant room for improvement remains in terms of accuracy. The GPT-2-RLLF-TT model exhibited a considerable improvement over GPT-2-ZS and GPT-2-TT but still remains suboptimal compared to state-of-the-art models. This gap may be addressed by further enhancing the models, adjusting the RLLF methodology, or exploring different LLMs. Additionally, the effectiveness of the Soft RLLF approach may vary depending on the domain and model size. Therefore, it is necessary to investigate its efficacy in other high-stakes domains and with larger models.

Moreover, one must consider the scalability of our approach when working with larger models like GPT-3 or GPT-4. Given the higher complexity and computational overhead associated with these models, developing cost-effective and efficient exploration-enhancement techniques becomes increasingly important. Further investigation and adaptation of our proposed approach to accommodate larger models would be necessary to maintain its effectiveness.

Despite the aforementioned limitations, our research provides valuable insights into the benefits of employing RLLF-enhanced exploration and transfer learning in LLMs. The improvements in negation understanding capabilities demonstrated by the GPT-2-RLLF-TT model have implications for the development of more accurate, reliable, and logically consistent language models in high-stakes domains that require precise negation understanding for robust decision-making processes.

\section{Conclusion}

In this paper, we presented a method for enhancing LLMs' negation understanding capabilities by balancing exploration and exploitation using Soft RLLF. Our approach employed the Ruletaker dataset for training a reward model, which was integrated into the GPT-2 RLLF training process to encourage the model to explore a broader range of negation possibilities. We also utilized transfer learning and evaluated the impact of our approach on the \xnot{} dataset in comparison to baseline models trained without RLLF.
The experimental results demonstrate that incorporating RLLF-enhanced exploration and transfer learning leads to substantial improvements in LLMs' negation understanding abilities. By enabling a more accurate understanding of negation, our proposed approach has the potential to significantly impact high-stakes domains such as law, healthcare.
Future research may further explore the limits of the Soft RLLF approach in different domains and with various model sizes, as well as investigate its scalability when working with larger and more complex models. Additionally, leveraging cost-effective and efficient exploration-enhancement techniques in tandem with Soft RLLF to optimize the performance of large-scale language models remains an area ripe for investigation.

\section*{Acknowledgement}
This work was supported by the AIP challenge funding related with JST, AIP Trilateral AI Research, Grant Number JPMJCR20G4.

\bibliographystyle{splncs04}
\bibliography{references.bib}
\end{document}